\documentclass[conference]{IEEEtran}
\IEEEoverridecommandlockouts
\usepackage{cite}
\usepackage{amsmath,amssymb,amsfonts}
\usepackage{algorithmic}
\usepackage{graphicx}
\usepackage{textcomp}
\usepackage{xcolor}
\usepackage{subcaption}
\usepackage{hyperref}
\def\BibTeX{{\rm B\kern-.05em{\sc i\kern-.025em b}\kern-.08em
    T\kern-.1667em\lower.7ex\hbox{E}\kern-.125emX}}
    \newcommand{\algname}{AWT }
    \RequirePackage{array}
  {\begingroup
   \newcommand\estyle{}%
   \renewcommand\institute[1]%
     {\\\multicolumn{#1}{@{}c@{}}{\scriptsize\begin{tabular}[t]{@{}>{\footnotesize}c@{}}##1\end{tabular}}}%
   \renewcommand\email[1]%
     {\gdef\estyle{\footnotesize\ttfamily}\\##1\gdef\estyle{}}
   \begin{tabular}[t]{@{}*{#1}{>{\estyle}c}@{}}
  }%
  {\end{tabular}%
   \endgroup
  }
\begin{document}

\title{AWT - Clustering Meteorological Time Series Using an Aggregated Wavelet Tree\\
{\small Extended version of the paper published at IEEE DSAA 2022}
\thanks{This
work was funded  in part  within the Austrian Climate and Research Programme ACRP project  KR19AC0K17614.}}


\author{\IEEEauthorblockN{Christina Pacher}
\IEEEauthorblockA{\textit{Data Mining and Machine Learning} \\
\textit{Faculty of Computer Science}\\
\textit{University of Vienna}\\
Vienna, Austria \\
christina.pacher@univie.ac.at}
\and
\IEEEauthorblockN{Irene Schicker}
\IEEEauthorblockA{\textit{Zentralanstalt f\"ur Meteorologie}\\
\textit{und Geodynamik} \\
Vienna, Austria \\
irene.schicker@zamg.ac.at}
\and
\IEEEauthorblockN{ Rosmarie DeWit}
\IEEEauthorblockA{\textit{Zentralanstalt f\"ur Meteorologie}\\
\textit{und Geodynamik} \\
Vienna, Austria \\
rosmarie.dewit@zamg.ac.at}

\and

%
\hspace{3.5cm}
\IEEEauthorblockN{Kate\v rina Hlav\'a\v ckov\'a-Schindler}
\IEEEauthorblockA{\hspace{3.5cm}\textit{Data Mining and Machine Learning} \\
\hspace{3.5cm}\textit{Faculty of Computer Science}\\
\hspace{3.5cm}\textit{University of Vienna}\\
\hspace{3.5cm}Vienna, Austria  \\
\hspace{3.5cm}katerina.schindlerova@univie.ac.at}
\and
\hspace{-2cm}
\IEEEauthorblockN{Claudia Plant}
\IEEEauthorblockA{\hspace{-2cm}\textit{Data Mining and Machine Learning} \\
\hspace{-2cm}\textit{Faculty of Computer Science}\\
\hspace{-2cm}\textit{University of Vienna}\\
\hspace{-2cm}Vienna, Austria  \\
\hspace{-2cm}claudia.plant@univie.ac.at}
}

%
%
%

\maketitle

\begin{abstract}
Both clustering and outlier detection play an important role for meteorological measurements. We present the \algname algorithm, a clustering algorithm for time series data that also performs implicit outlier detection during the clustering. \algname integrates ideas of several well-known K-Means clustering algorithms. It chooses the number of clusters automatically based on a user-defined threshold parameter, and it can be used for heterogeneous meteorological input data as well as for data sets that exceed the available memory size. We apply AWT to crowd sourced 2-m temperature data with an hourly resolution from the city of Vienna to detect outliers and to investigate if the final clusters show general similarities and similarities with urban land-use characteristics. It is shown that both the outlier detection and the implicit mapping to land-use characteristic is possible with AWT  which  opens new possible fields of application, specifically in the rapidly evolving field of urban climate and urban weather.
\end{abstract}

\begin{IEEEkeywords}
Time Series Clustering, Outlier Detection, Meteorological Data, BIRCH, I-Kmeans
\end{IEEEkeywords}

\section{Introduction}
 In the past years the amount of acquired data  in meteorology and climatology increased considerable especially with the increasing availability of Internet  access and connected structures and the usage of rather unusual data for meteorological applications (e.g. microlinks). These unusual connected objects have gained interest as they provide additional, more fine grained information on the state of the lower atmosphere. They are, however, prone to data quality issues and, due to the large amount of such supplementary data, the increasing high dimensionality of meteorological data. This poses a challenge in terms of quality control before using them as they add an additional time lag factor before  they can be used. To apply a quality control method to detect the transitions, outliers, and relationships, especially with non-standard data, methods able to work with high dimensional time series data are thus required.

One technique for identifying such underlying structures in exploratory data analysis, quality control and feature extraction are clustering algorithms \cite{halkidi2001}. In clustering, the objects sharing similar characteristics are grouped by using different distance metrics, enabling users to extract information.  This information might have even been unnoticed by the experienced user. As clustering and outlier detection are closely related tasks, one can use clustering results as a first quality control step for identifying outliers in data. In order to find outliers, the algorithm/user must learn what  regular patterns exists in the data, corresponding typically to the large clusters. Outliers are thus those data objects that do not fit well into any of these large clusters or are, in terms of clustering e.g. meteorological ensemble forecasts, those that are very unlikely to occur and represent upper/lower tails of distributions of the data.

Clustering and outlier detection has been widely applied in meteorology and climatology in many applications  in the past decades but especially in clustering ensemble predictions such as the ECMWF ensemble prediction or the characterization of spatial and temporal patterns of rainfall.  Methods used and commonly applied are the K-Means algorithm, decomposition, fuzzy-c-means, self organising maps (SOM), hidden Markov Models, and others. K-Means has been applied by \cite{casellas2020} to cluster observation sites for the interpolation of temperature and humidity measurements from point to grid for Catalunya.  \cite{Zarnani2014}  implemented different (crips) clustering algorithms, namely K-Means, CLARA, and hierarchical clustering, as well as Fuzzy C-means to forecast prediction interval uncertainties using NWP forecasts. \cite{Sathiaraj2018} applied K-Means as well as two clustering methods, DBSCAN and BIRCH, to demonstrate the skills of each method in predicting the climate types for the continental United States.  An extensive literature review on clustering   application to the characterisation of air pollution patterns is given in \cite{GOVENDER202040}. Other applications range from  the estimation of ensemble model uncertainty \cite{clustering-mediumrange}, the detection of extreme event patterns \cite{Seibertetal2007}, air quality  \cite{defoy2008, DAVIS1998}, characterization of spatial and temporal patterns of rainfall \cite{Lyraetal2014} or for data grouping in forecasting of wind speed for wind turbines \cite{papazek2020}.  For traditional meteorological data clustering is often used in pre- and post-processing.

 With the increasing availability and usage of non-traditional meteorological data sets such as privately owned weather sites (PWS) of different companies a new source of information in the lower atmosphere is available. This information can be used in data assimilation of numerical weather prediction models (NWP) or in the generation e.g. of gridded temperature fields for the verification of urban climate models. Additionally, as the amount of PWS is typically higher in urbanized areas, these data can give insights into effects such as the urban heat island (UHI), e.g. \cite{Oke1995}. The rather dense installation of such PWS sites allows to gain additional information as meteorological providers do not operate this densely. One of the companies selling such outdoor weather station modules is NetAtmo. If the user agrees, the collected meteorological data is published online automatically, and this data can be then freely downloaded.  However, the user also has to agree to share certain information such as location to enable NetAtmo to properly display the data geographically. If the user does not agree, the site is assigned to  the town/village center.  Still, these collected information yield an incredible source of additional information on the state of the local atmosphere. As a logical consequence, \cite{Chapman2017} proposed to use  these stations to supplement the professional observational networks operated by e.g. weather services.
  However, those PWS data have no strict quality rules in terms of siting, ventilation, and more. Thus,  the data quality of the observations recorded by these smart home devices is not known. This is in contrast to data from professional networks, which follow high quality standards \cite{wmo2018}. To be able to use PWS data one needs to implement and apply  a stringent quality control prior to further use. The data quality may be compromised by a variety of factors, including sensor design (e.g. non-ventilated and sealed temperature sensors), calibration and metadata issues (e.g. no location information specified, wrong altitude), and software and communication problems as well as unsuitable installation locations, e.g. \cite{Chapman2017},
  \cite{Meier2017}.

 More recently, work has been carried out to develop quality control algorithms for PWS sites.\cite{Meier2017} noted that the main error sources of privately owned weather stations are related to the siting of the sensors rather than sensor quality itself, and developed a quality control algorithm that specifically addresses errors resulting from a disadvantageous sensor installation. Based on their procedure,  \cite{Napoly2018} developed an automated statistical quality control. In contrast to the \cite{Meier2017} method, their method does not rely on reference data from a professional meteorological network, allowing the method to be applied in cities without such a network. Slight modifications of the above introduced quality control algorithms have successfully been used by e.g. \cite{Hammerberg2018} and \cite{Feichtinger2020} to analyse the spatial distribution of temperatures in Vienna, Austria, with the idea to detect the diversity in land surface of urban areas.

The approaches based on the work by \cite{Meier2017} are, however, quite restrictive and  stations with too many outliers per month are removed by the quality control algorithm. Stations with faulty measurements during the daytime as a result of radiative errors are removed for the entire time period under consideration, meaning that potentially high-quality night-time observations are discarded as well. One way to solve this issue is by using pattern recognition or clustering techniques focusing at single time steps to sort the observations, rather than at time series. One such approach is described in \cite{Nipen2020}, who used data from the NetAtmo network to improve the operational nowcasting in the Greater Oslo area, Norway, The TITAN (auTomatIc daTa quAlity coNtrol) package. In TITAN, every hour is quality controlled independently following different steps: a spatiality check is carried out consisting of a buddy check, a spatial consistency check, and a loneliness test. This quality controlled data is then pre-processed to enable  numerical weather prediction or urban/climate models to assimilate the data. \cite{Mandement2020} and \cite{Caumont2021} used personal weather sites (PWS) to investigate deep convection and heavy precipitation events in France. They developed and applied a quality control algorithm as well as a comparison to the observation sites and highlighted the additional value of such data.

In this paper we present a novel clustering algorithm for meteorological data that also enables outlier detection without knowledge of physics,  climate, environment of the surroundings, or reference stations. Thus, no buddy check or radiation correction is needed, which makes the clustering algorithm applicable also for data where hardly any standard measurements are taken, such as at hub height of wind turbines. In contrast to algorithms such as K-Means the proposed novel algorithm, AWT (Clustering using an Aggregated Wavelet Tree), a modification and combination of several well-known clustering algorithms, does not require the number of clusters to be specified as an input parameter. Similar to the classical BIRCH algorithm, our method AWT works on a multi-resolution data structure, an Aggregated Wavelet Tree that is suitable for representing multivariate time series. In contrast to K-Means and BIRCH, the user does not need to specify the number of clusters K, which is often difficult in many applications. Instead, AWT relies on a single threshold parameter for clustering and outlier detection. This threshold corresponds to the highest resolution of the tree. Points that are not in any cluster with respect to the threshold are naturally flagged as outliers. Moreover, the proposed  algorithm is able to work efficiently with time series, multiple time series per site, and is robust in the presences of outliers.

The paper is structured as follows: The clustering algorithm is described in Section \ref{sec:algorithm},  its application to the NetATMO sites located in Vienna is shown in Section \ref{sec:experiments}, and Section \ref{sec:conclusion} concludes the results.

\section{Clustering Algorithm} \label{sec:algorithm}
In its basic idea, the \algname algorithm is similar to the approach presented by \cite{BIRCH+IKmeans}, which combine the BIRCH \cite{BIRCH} and I-Kmeans \cite{IKmeans} algorithms. However, we introduce a few significant modifications leading to a thorough reinterpretation of how the clustering problem is approached.
Section \ref{alg:basis}  first reviews the well-known clustering algorithms on which our approach is based. Then,  we  describe in Section \ref{alg:description} how the \algname algorithm is derived from them, and examine its properties and possible extensions.

\subsection{Basis Algorithms} \label{alg:basis}

\subsubsection{BIRCH: Balanced Iterative Reducing and Clustering using Hierarchies}

BIRCH \cite{BIRCH} is a hierarchical clustering algorithm originally designed to work with data sets that contain too many data points to fit into the main memory. It is well suited for this task as it stores a compressed representation of the clustering instead of the full data, and it only requires one pass over the data set to achieve a good clustering result.
BIRCH stores the information about the clustering it creates as a tree, where each node represents a cluster. It is hierarchical in the sense that the cluster represented by any node is the union of all the clusters represented by this node's children. The root of the tree therefore represents one big cluster that contains the whole data set.
Information about a cluster is compressed into the so-called \emph{clustering feature}, a triple $(N, \vec{LS}, SS)$.
$N$ is the number of data points in the cluster;
$\vec{LS}$ is the linear sum of all data points $\vec{X_i}$ in the cluster, i.e. $\vec{LS} = \sum_{i=1}^N \vec{X_i}$;
and $SS = \sum_{i=1}^N \vec{X_i}^2$ is the square sum of all data points $\vec{X_i}$ in the cluster \cite{BIRCH}. The square of $\vec{X_i}$ is the inner product of $\vec{X_i}$ with itself.
This information is sufficient to compute statistical measures about clusters or merge two clusters together. Therefore, all the information that has to be stored at a node to accurately represent a cluster is the clustering feature of this cluster -- it is not necessary to store the individual data points. For this reason, BIRCH's tree representation of the clustering is called the \emph{clustering feature tree}.

To insert a new data point into a cluster, the tree is traversed to find the leaf node that is \emph{closest} to the new point. This can simply be done by starting at the root and recursively choosing the child cluster of the current node that is closest to the new point, until a leaf node is reached. The distance between a cluster and a data point is defined as the distance between the data point and the cluster's centroid. Using the clustering feature, the centroid can be computed as $\vec{LS}/N$.
Once the closest leaf has been found, there are two options: Either the new data point will be merged into this leaf cluster, or a new leaf cluster will be created that contains the new data point. To decide which option is chosen, BIRCH uses a \emph{threshold}, which is, simply put, an upper limit for how far away from each other points in the same leaf cluster are allowed to be. This requires the use of some measure for the \emph{tightness} of a cluster, to evaluate the tightness of the cluster that would be created by merging the new point into the leaf. If the result is less than the threshold, the point will be merged. Otherwise, a new leaf cluster will be created that only contains the new data point. In both cases, all nodes on the path to the root have to be updated to reflect the new state.

\begin{figure}
    \centering
   \includegraphics[width=0.42\textwidth,
   clip]{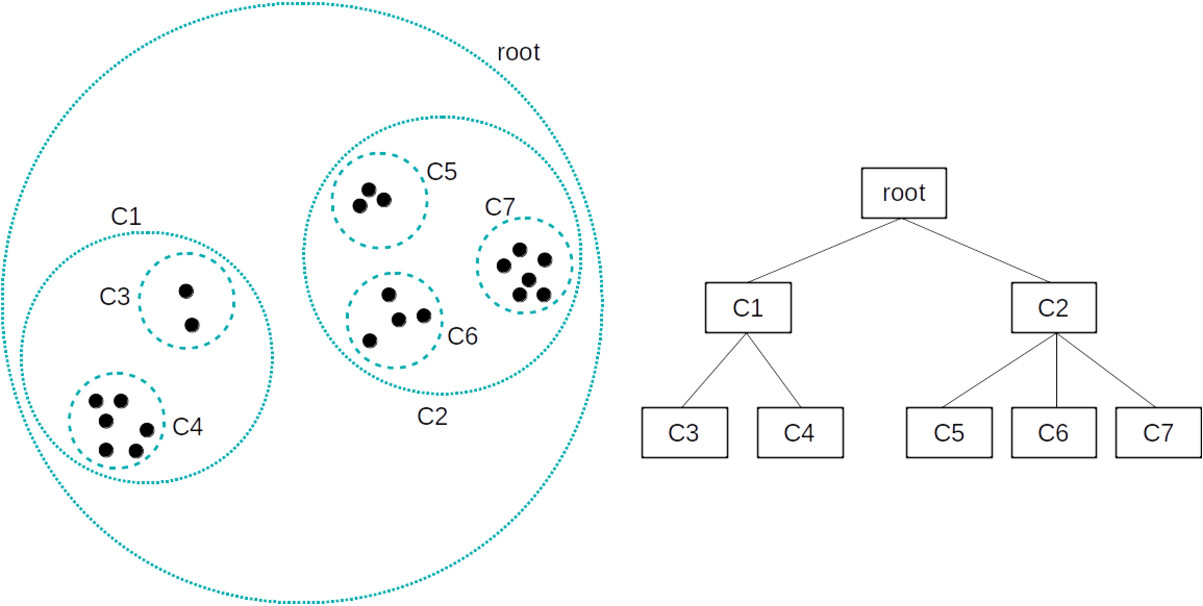}
    \caption{An example of the clustering hierarchy represented by a clustering feature tree. The size of the leaf level clusters (C3 -- C7) is determined by the threshold.}
    \label{fig:cftree}
\end{figure}

Fig. \ref{fig:cftree} shows an example for a clustering feature tree as it is created by the BIRCH algorithm.
The tightness criterion used in this article is the \emph{average inter-cluster distance} between two clusters (the new data point is considered to be a cluster of its own). For two clusters $C_1, C_2$, the average inter-cluster distance is defined as follows:

\begin{equation}
D(C_1,C_2) = \sqrt{
\frac{
\sum_{i=1}^{N_1} \sum_{j=1}^{N_2} (\vec{X_i} - \vec{Y_j})^2
}{N_1*N_2}
}
\end{equation}

If $(N_1, \vec{LS_1}, SS_1)$ and $(N_2, \vec{LS_2}, SS_2)$ are the clustering features of $C_1$ and $C_2$, respectively, this is equivalent to:

\begin{equation}
D(C_1, C_2) = \sqrt{\frac{N_1*SS_2 + N_2*SS_1 - 2\vec{LS_1}\vec{LS_2}}{N_1*N_2}}
\end{equation}
To reduce computational effort, our algorithm actually uses the \emph{squared} average inter-cluster distance, i.e. it omits the computation of the square root. This can easily be done because we do not care about the exact results, but only about their proportions. For similar reasons, when computing the distance between a data point and a cluster centroid we use the squared Euclidean distance.

BIRCH has a \emph{branchingfactor} parameter that defines the maximum number of children that any non-leaf node may have. If inserting a new leaf cluster exceeds this branching factor, this causes a node split. Potentially, this node split might propagate all the way to the root, in which case the height of the tree increases.
One key idea with BIRCH is that the granularity of the leaf level clusters should be as fine as possible while still allowing the whole clustering feature tree to fit into the main memory. To this end the threshold parameter can be adapted.
This leads to a trade-off between the granularity of the clustering result and the required amount of memory:
The larger the threshold, the more coarse-grained the clustering will be, but the less memory it will require. The threshold should therefore be as small as possible within the memory constraints.

After the clustering feature tree has been created, the clustering can be improved further by applying any in-memory clustering algorithm to the leaf-level clusters -- i.e. treating each leaf-level cluster as a single data point. Obviously, the finer the granularity of the leaf-level clusters, the more exact the result of this final clustering phase will be. The parameters of the chosen in-memory clustering algorithm can be set by the user to produce the desired number of output clusters. \cite{BIRCH} recommend the use of an in-memory clustering algorithm because, while the quality of the clustering represented by the clustering feature tree is usually already quite good, the result does depend on the input order of the data points and can still be considerably improved by applying a few additional in-memory clustering iterations.
 In this study we use the BIRCH algorithm for comparison with our AWT algorithm.

\subsubsection{I-Kmeans}

\begin{figure}
\centering
\begin{subfigure}[t]{.22\textwidth}
  \centering
  \includegraphics[width=\linewidth]{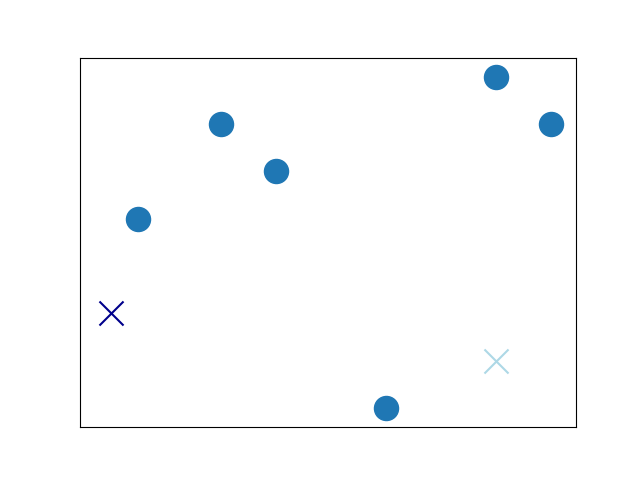}
  \caption{}
  \label{fig:kmeans:1}
\end{subfigure}%
\begin{subfigure}[t]{.22\textwidth}
  \centering
  \includegraphics[width=\linewidth]{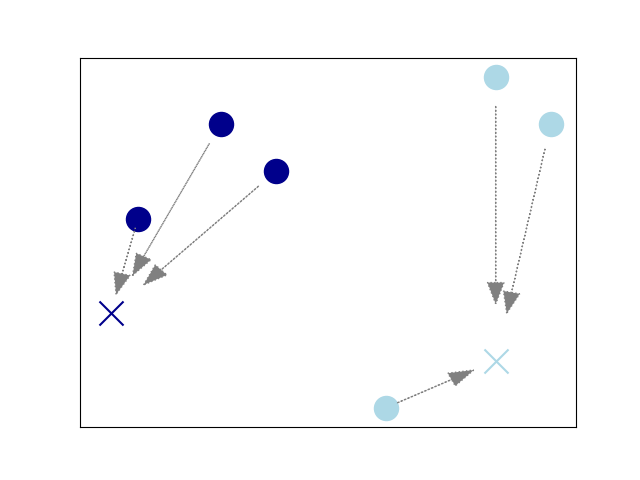}
   \caption{}
  \label{fig:kmeans:2}
\end{subfigure}%
\\
\begin{subfigure}[t]{.22\textwidth}
  \centering
  \includegraphics[width=\linewidth]{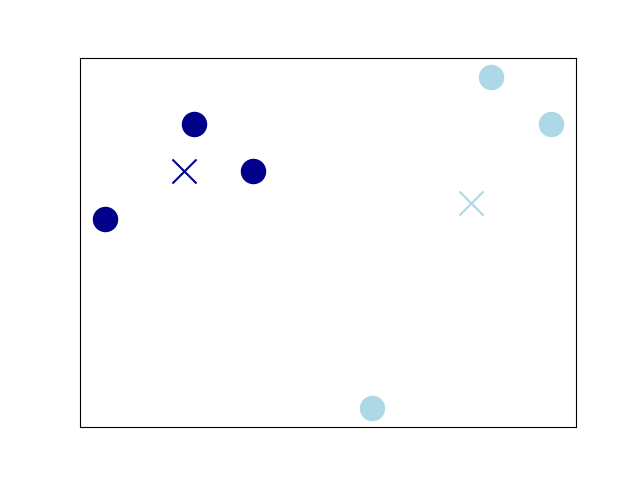}
   \caption{}
  \label{fig:kmeans:3}
\end{subfigure}
\caption{ An example for  K-Means algorithm with $k=2$ (data points are  dots, cluster centroids are  crosses). (a) During initialization, a centroid is chosen for each cluster (e.g. randomly); Steps (b) (i.e. each data point is assigned to the closest centroid) and (c) (i.e. the cluster centroids are recomputed) are repeated until the algorithm converges, i.e. until all points remain in the same cluster during the reassignment step (b).}
\label{fig:kmeans}
\end{figure}

\begin{figure}
    \centering
    \includegraphics[width=0.4\textwidth]{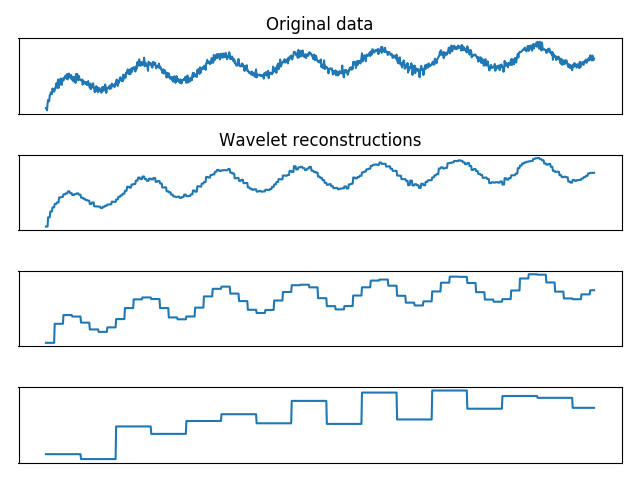}
    \caption{A (synthetic) time series and various wavelet approximations, using the Haar wavelet. The number of wavelet levels used for the approximation decreases from top to bottom. The more wavelet levels, the closer is the result is to the original data, but the important trends in the data are already captured by the coarse wavelet approximations.}
    \label{fig:wavelets}
\end{figure}

I-Kmeans (\cite{IKmeans}) is a variation of the well-known K-Means algorithm (\cite{Lloyd_kmeans}, \cite{McQueen_kmeans}) that has been adapted to work with time series. The basic idea is the same as with the "ordinary" K-Means. The user defines the number of clusters that should be created (the $k$ parameter). An initial centroid is chosen for each cluster through some initialization strategy (random initialization in the simplest case). In each iteration of the algorithm, each data point is assigned to its closest cluster -- the cluster where the distance between the point and the centroid is the smallest -- and then the cluster centroids are recomputed. The algorithm converges when no point changes its assigned cluster. The basic K-Means algorithm is visualized in Fig. \ref{fig:kmeans}.
To make this algorithm suitable for time series data, I-Kmeans works on a wavelet representation of the timelines. Wavelets have different resolution levels, where the lower levels provide a coarse approximation to the real data. The higher the resolution is, the more detailed this approximation becomes, until the full resolution gives an exact representation of the original data. An example for this is shown in Fig. \ref{fig:wavelets}.
\cite{IKmeans} also provide a helpful visualization of wavelet approximation specifically for the Haar wavelet, including a visualization of the wavelet basis functions. A more in-depth introduction to wavelets in general can be found for example in \cite{WaveletIntroduction}.

I-Kmeans starts by clustering low-resolution representations of the data points and increases the resolution in each subsequent iteration. For our algorithm we use a Haar wavelet decomposition, which is also done by \cite{IKmeans}; however, any decomposition that has the described multi-resolution property could be used instead.
Clustering wavelets has several advantages. Using low-resolution approximations at the beginning of the iteration process means that the essential trends in the data are captured, but finer details are ignored, which reduces the likelihood of getting trapped in local optima. The reduced dimensionality of the low-resolution wavelets also means that the work required for the distance computations is significantly reduced during the first iterations. \cite{IKmeans} show that I-Kmeans outperforms the ordinary K-Means algorithm both in terms of runtime and quality of the result, and that it often converges already at low resolutions. In fact, it is not even necessary to store the full wavelet resolution: High quality clustering results can still be achieved if the highest resolution levels are discarded and only the remaining levels are used for clustering.\footnote{This is especially effective in terms of memory usage since the number of coefficients per wavelet level increases exponentially. Dropping only the highest resolution level will already cut the memory requirements in half.}

There are two important problems generally associated with getting a good result from any K-Means variant: the choice of the $k$ parameter and of the initial centroids. If the chosen number of clusters does not match the natural structure of the data, the resulting clustering cannot reflect the actual clusters present in the data; and even if the chosen number of clusters is correct, a poor initial choice of centroids can lead to a poor result.
\cite{BIRCH+IKmeans} propose a combination of BIRCH and I-Kmeans.
At first, the clustering feature tree is built with wavelet approximations instead of the raw data. I-Kmeans is then used as the in-memory clustering algorithm that is applied to the leaf-level clusters during the final phase of BIRCH.
This modification introduces all of the advantages of clustering wavelets into the BIRCH algorithm and thus makes BIRCH suitable for time series data.
Moreover, the aforementioned problem of choosing the initial cluster centers for I-Kmeans is resolved: The initial guess for the centroids can be derived from the clustering feature tree.

\subsection{The \algname Algorithm} \label{alg:description}

The basic idea of \algname is very similar to the algorithm described by \cite{BIRCH+IKmeans}: A clustering feature tree is created from wavelet approximations of the time series data, and then I-Kmeans is used to improve the clustering result. However, we introduced some important changes, which make a considerable difference in the functionality of the algorithm. These changes will be described here.

\subsubsection{Reinterpreting the Threshold Parameter} \label{alg:threshold}

As mentioned before, BIRCH was intended to work with data sets that do not fit into the main memory due to the large number of data points. However, data that results from meteorological measurements is usually "tall and thin": The number of weather station is usually limited, and depending on the examined time period there might well be thousands of measurements per station. In other words, there is a comparatively small number of data points, and each data point has a large dimensionality. For this kind of data, we can therefore safely assume that if the data set exceeds the available memory, this will be due to the number of dimensions per data point, and not due to the number of data points.
Furthermore, since we are working with time series, it is possible to use wavelet representations as described above; and, as \cite{IKmeans} point out, it is not actually necessary to store the full wavelet resolution, since the I-Kmeans algorithm already provides good results at low wavelet resolution levels. As a consequence, we can assume that it is possible to make the data set fit into the main memory by applying a wavelet transformation to each time series and discarding an appropriate number of highest wavelet resolution levels.

\begin{figure}
\centering
\begin{subfigure}{.22\textwidth}
  \centering
  \includegraphics[height=3cm]{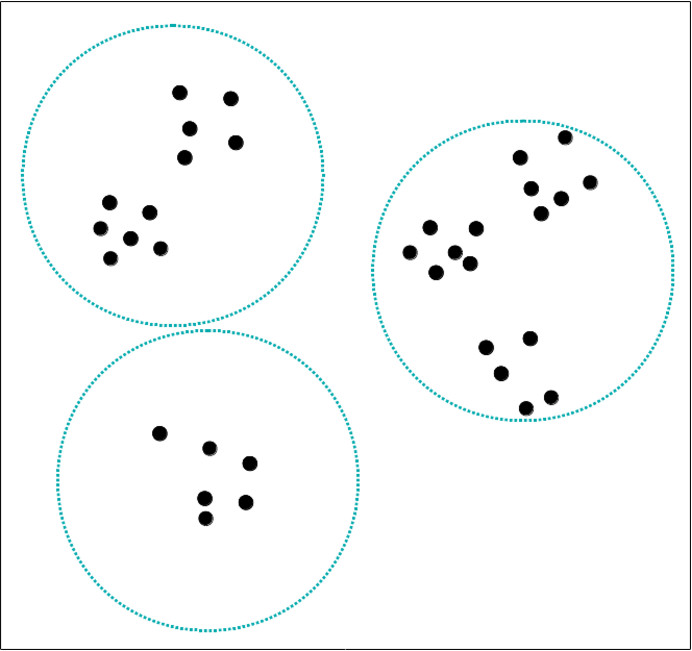}
  \caption{}
  \label{fig:thresh:large}
\end{subfigure}%
\begin{subfigure}{.22\textwidth}
  \centering
  \includegraphics[height=3cm]{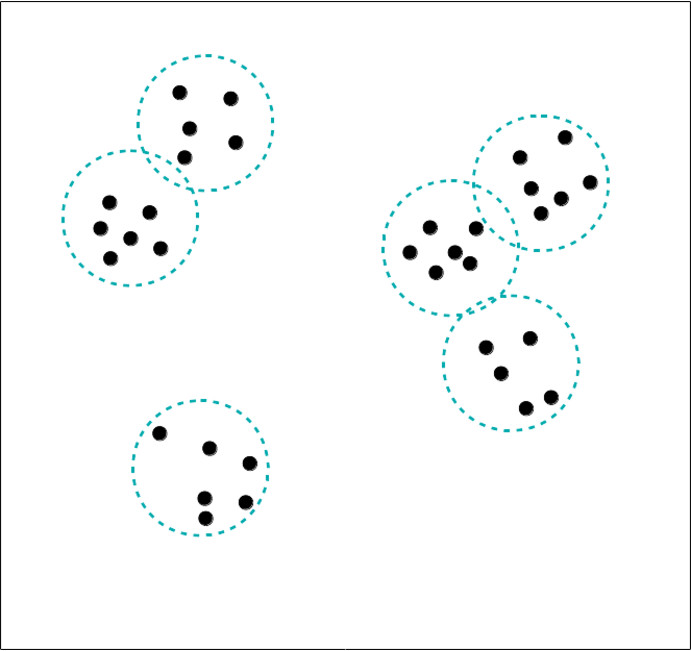}
  \caption{}
  \label{fig:thresh:small}
\end{subfigure}
\caption{How the threshold influences the size and number of clusters in AWT. (a) A large threshold creates a few large clusters. (b) A small threshold creates many small clusters.}
\label{fig:thresh}
\end{figure}

If we accept these assumptions, this means that we no longer need to use the threshold to make the clustering feature tree fit into main memory. Instead, we consider the threshold to represent the desired tightness of the output clusters. In other words, we consider the leaf-level clusters to represent the desired granularity of the result.
Increasing the threshold will lead to fewer, larger clusters, while using a smaller threshold will lead to many small clusters. Since the exact number of clusters depends on the data, this algorithm is very likely to produce clusters that exist naturally in the data (contrary to the standard K-Means, where a poor choice of $k$ can lead to a clustering result that does not represent the natural structure of the data at all). An example for how the clustering result is influenced by different threshold values is shown in Fig. \ref{fig:thresh}.

It should be pointed out that the distance between two wavelets depends on the current wavelet resolution: The more wavelet coefficients are taken into account for the distance computation, the larger the distance will be. This also affects the computation of the tightness of a cluster: Here, the average inter-cluster distance is affected by the number of wavelet levels that are used to build the clustering feature tree. This number can be chosen by the user. As a consequence, if the threshold stays the same, but the number of wavelet levels in the clustering feature tree is changed, the resulting number of clusters will also change.

When combining this new BIRCH-variation and I-Kmeans, we can not only derive good initial cluster centers for I-Kmeans from the clustering feature tree, but we also derive the $k$ parameter for I-Kmeans by simply counting the leaf nodes in the tree. Thus our reinterpretation of the threshold resolves both major challenges associated with K-Means variations.

\subsubsection{Extension to Panel Data}

It may be desirable to use more than one meteorological parameter as clustering input -- for example, we might want our algorithm to consider both the air temperature and the air pressure at a weather station. Extending the algorithms described above to this situation is straightforward and only requires an appropriate distance measure for data that contains multiple time series per point (so-called \emph{panel data}).

\cite{paneldata} propose such a distance function. Note that when using this approach, all time series should be scaled during preprocessing. This ensures that all meteorological parameters contribute equally to the result, despite their different units of measurement and different possible value ranges.

To compute the distance between two points, the distance measure is at first computed for each timeline individually. In our example, one would compute one value to measure the distance between the air temperature measurements at the two points, and another value to measure the distance between their air pressure measurements.
The final distance measure is then computed by simply summing up all of these individual distances.
For the \algname algorithm, we adapt this function to work on wavelet data. This is achieved by only using the coefficients up to the current wavelet resolution level for the distance computations.

To formalize: Let $n$ be the number of time series per point (i.e. the number of different meteorological parameters measured at each location), and let $r$ be the total number of wavelet coefficients at all levels up to and including the current wavelet resolution level. Then the distance between two data points $X$ and $Y$ is defined as:

\begin{equation}
dist(X,Y) = \sum_{i=1}^n \sum_{j=1}^r (X[i][j] - Y[i][j])^2
\end{equation}

\subsubsection{Outlier Detection with AWT}

\begin{figure}
\centering
\begin{subfigure}{.22\textwidth}
  \centering
  \includegraphics[height=3cm]{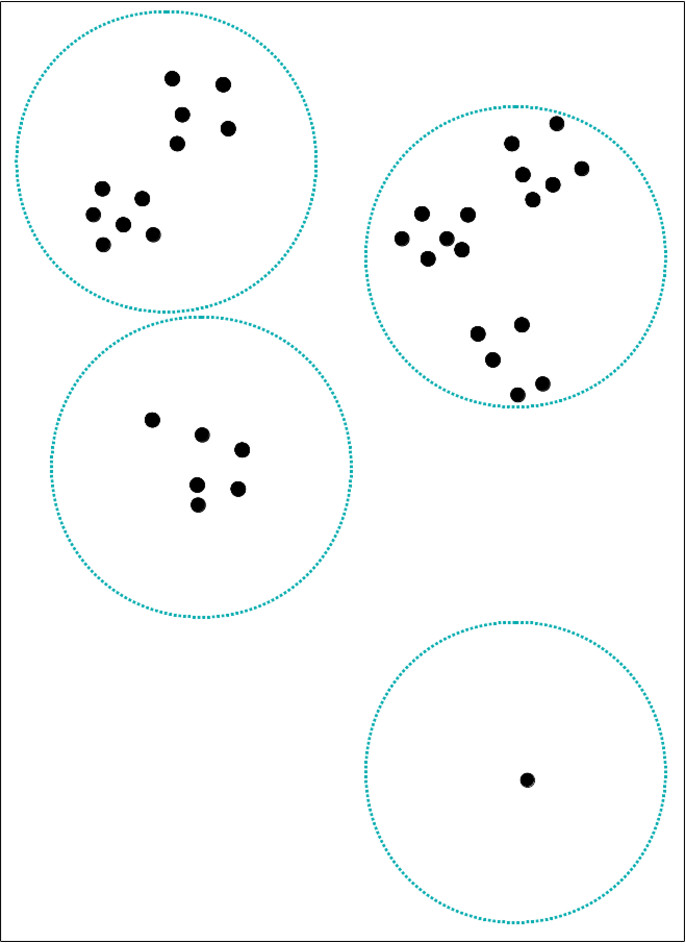}
\end{subfigure}%
\begin{subfigure}{.22\textwidth}
  \centering
  \includegraphics[height=3cm]{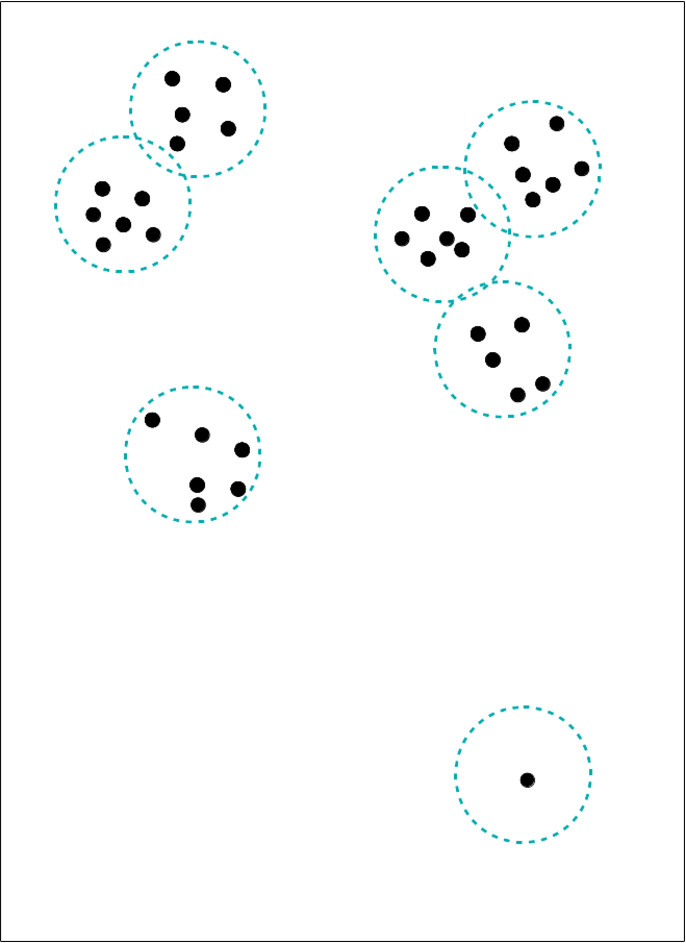}
\end{subfigure}
\caption{\algname in the presence of outliers, for a small and a large threshold. In both cases the outlier data point ends up in a cluster of its own.}
\label{fig:thresh:outlier}
\end{figure}

A very useful property of \algname is that it will not be disturbed by the presence of outliers in the data and is even able to identify them as a natural part of the clustering process, without requiring any additional work at all. This is again due to the functionality of the threshold value.
If a data point is very different from every other point in the data set (i.e. it is an outlier), then merging it into a common cluster with any other data point would always cause the resulting cluster to exceed the threshold, and so the outlier will end up in a cluster of its own. Therefore, outlier detection can be performed by simply labelling all clusters that contain only one point (or possibly a very small number of points) as outliers. An example for this behaviour is shown in Fig. \ref{fig:thresh:outlier}.

Obviously, if the threshold is extremely large or extremely small, this outlier detection functionality will no longer work. A very large threshold will eventually lead to a situation where the outlier points may be merged into existing clusters after all. On the other hand, if the threshold is too small, hardly any points will be allowed to merge into an existing cluster, and the result will look as if the data set consists almost entirely of outliers. However, in these cases the clustering itself would not be very valuable any more, as it would be either too coarse or too fine to really show any significant information about the structure in the data. In the usual case where the threshold is not chosen extremely large or extremely small, we can expect the result to show the structure of the data and to correctly identify outliers.

\section{Experiments} \label{sec:experiments}
We applied our algorithm to data from NetAtmo stations in the area of Vienna, Austria and evaluated it against the BIRCH algorithm as well as against measurements from the Austrian meteorological service ZAMG. For our experiments we chose measurements from a six-month time window from September 2018 to February 2019, taken at 1-hour intervals. This corresponds to the meteorological seasons autumn (September -- November) and winter (December -- February). In all of our experiments we used air temperature measurements only.

This data set is small enough to fit into the main memory as a whole, without the need to reduce the wavelet resolution first. This will in turn allow us to perform a comparison between the clustering results at full resolution and the clustering results at reduced resolution, and to examine how much the result is impacted by the reduction of the wavelet resolution.

Unless explicitly stated otherwise, all results described in this Section arise from experiments with a threshold value of $1.0$, with three wavelet levels used to create the clustering feature tree. At full wavelet resolution, the data contains 13 wavelet levels. Since no reduction of the amount of data was necessary with the size of this data set, no wavelet levels were discarded  for this set of experiments.
The result contains 46 clusters.

\subsection{Data Preprocessing}
When dealing with crowd sourced data, there are some points to be taken into consideration that do not arise with data from officially managed weather stations.
The fact that there is no centralized management of the NetAtmo stations implies that there is no quality control. Incorrect or improper installation can lead to erroneous measurement results, such as radiation errors if a station is set up on a balcony facing south resulting in air temperature measurements that are generally too high during the day.
Furthermore, the record time stamps of the measurements are not synchronized among stations. Thus, in contrast to standard meteorological sites here one needs to consider a time window of several minutes when sampling data from multiple stations or for comparisons/evaluation against standard meteorological sites. Another factor which needs to be taken into account is the operation time of such privately owned sites. Weather stations / modules may join or leave the "crowd", i.e. stop recording at a specific location, at any moment due to different reasons (low battery, relocation, privacy settings of the owners, etc.). There is, thus, a strong possibility that stations lack recordings when considering data from the comparatively long time span of six months.

In spite of the expected quality issues, we did not perform any actual quality control beforehand, in the sense that we did not verify if the reported measurements were realistic. Instead, we expect the \algname algorithm to be able to deal with unrealistic values by simply labelling them as outliers. In contrast to other algorithms, no additional climate consistency check is performed as we expect that the data are additionally quality controlled by any further model in the application line. However, there were a few other preprocessing steps that had to be taken.

Because of the high probability that data entries might be missing from the database, we needed a way to react to missing data items. We specified a tolerance of 10 percent for missing data: if less than 10 percent of a station's measurements are missing, these values are reconstructed using linear interpolation, otherwise the station is excluded from the data set.
For the linear interpolation, we chose a very simple approach that only takes into account the values from the same station.
\footnote{This approach will lead to very unrealistic results in cases where data is missing at the beginning or end of the time series, as the values created by the interpolation method will simply be copies of the closest existing value. Rather than using a more sophisticated preprocessing method, we accepted this possibility because we expect the \algname algorithm to simply flag these unrealistic time series as outliers.}

Vienna lies in a relatively flat region compared to other parts of Austria, but still there is a height difference of almost 400 meters between its highest and lowest point (cf. \cite{wien-stadtgebiet}).
Since air temperature decreases with increasing height at a rate of about $0.65K / 100m$, this height difference might be enough to distort the clustering result.  Having in mind a usage of this algorithm for domains covering larger regions with a higher variance in altitude either using point or panel data, a height correction is necessary.
We therefore applied height correction to filter out temperature differences that are caused only by the difference in height between the stations. This height correction is performed using the following formula:

\begin{equation}
t_{corr} = t + ((0.0065 * (z - mz))
\end{equation}

where $t_{corr}$ is the corrected temperature, $t$ is the original temperature measurement, $z$ is the height of the station where the measurement was taken, and $mz$ is the mean height of all stations in the data set.

For the purposes of this height correction, we assumed that all stations were installed at ground level. This is quite probably not the case, but given the notorious unreliability of height information available for the NetAtmo stations, it is a reasonable simplification. Influences of the different types of urban and land cover are seen in the clustering results (see Section Clustering Results). 
In addition, we excluded all stations from the data set that did not have proper coordinate information available. Plotting these stations on the map or even applying height correction to them would neither be feasible nor yield reliable results, and by excluding these stations we ensure that proper analysis and visualization of our clustering results is possible.
While these preprocessing steps are rather basic, they nevertheless led to a huge diminution of the size of the data set: Of the approximately 1400 stations that provide measurements for our chosen six-month time window, only 608 remain available for clustering.

 To put the height corrected and pre-processed NetAtmo measurements in relation to observations of WMO-standard measurements, data of the Austrian TeilAutomatische WetterStation (TAWES), the semi-automatic national weather service observation sites, are used. Here, data of eight measurement locations inside Vienna are used. Fig. 6 
 (left) shows an overview of the location of the TAWES sites (blue triangles) as well as the location of the used NetAtmo sites (pink dots). In Fig. 6 
 (right) the temperature measurements within the period considered is shown. The black and thin lines show the more than 600 NetAtmo sites, the magenta / linespoints line the temperature average measured by the eight TAWES sites in Vienna. What can be seen from a visual inspection is that besides on average higher temperatures at the NetAtmo sites some clearly distinguishable outliers are present. These are both outliers in terms of missing seasonal cycle (and very likely measurement devices which are placed indoors), and in terms of a too high diurnal amplitude and very high temperature values. However, there are also outliers which can not be seen from visual inspection only as will be seen in the following subsections.

\begin{figure}
\begin{subfigure}[t]{.25\textwidth}
  \centering
    \includegraphics[width=0.99\textwidth]{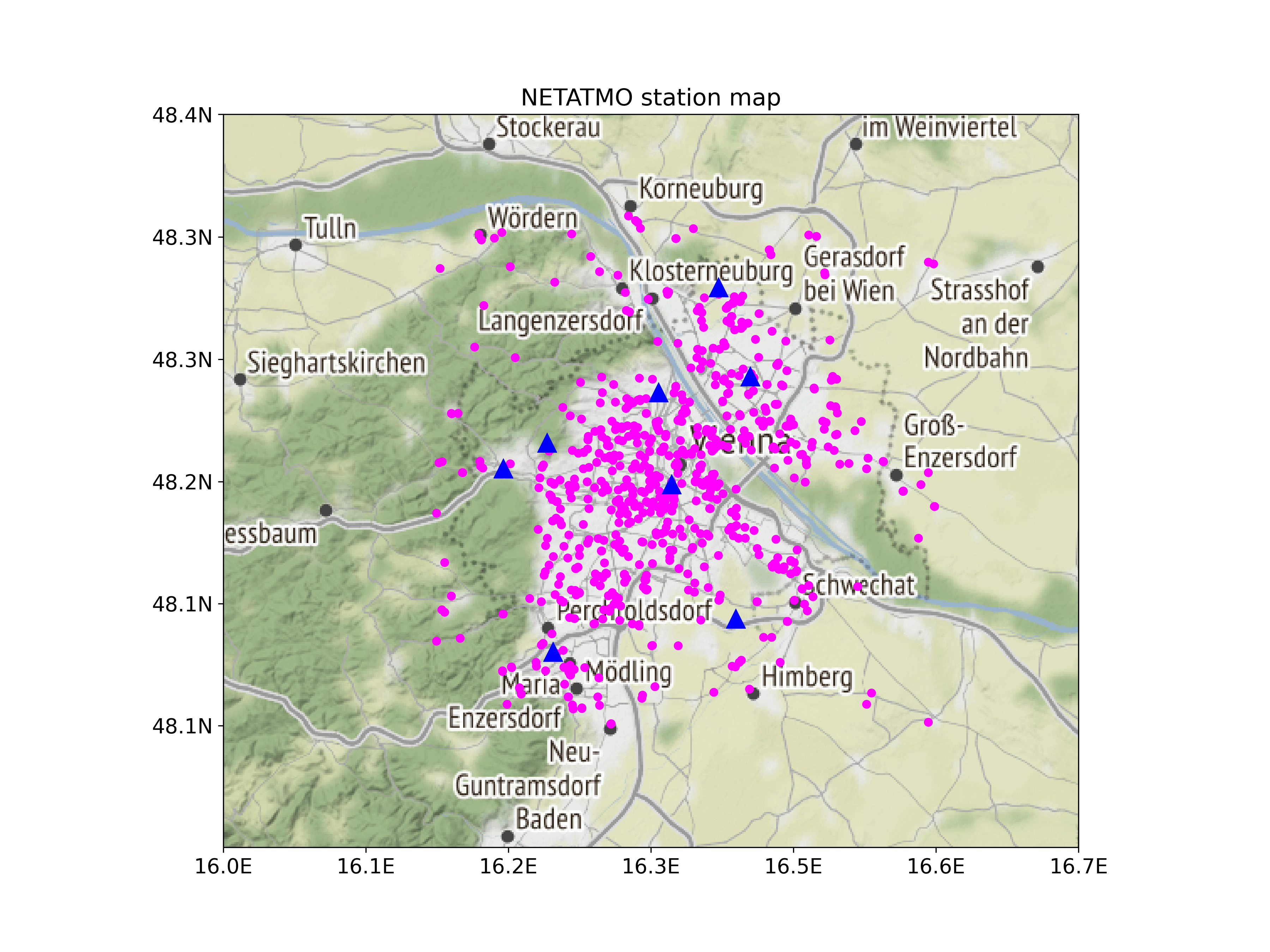}
\end{subfigure}%
\begin{subfigure}[t]{.25\textwidth}
  \centering
  \includegraphics[width=0.99\linewidth]{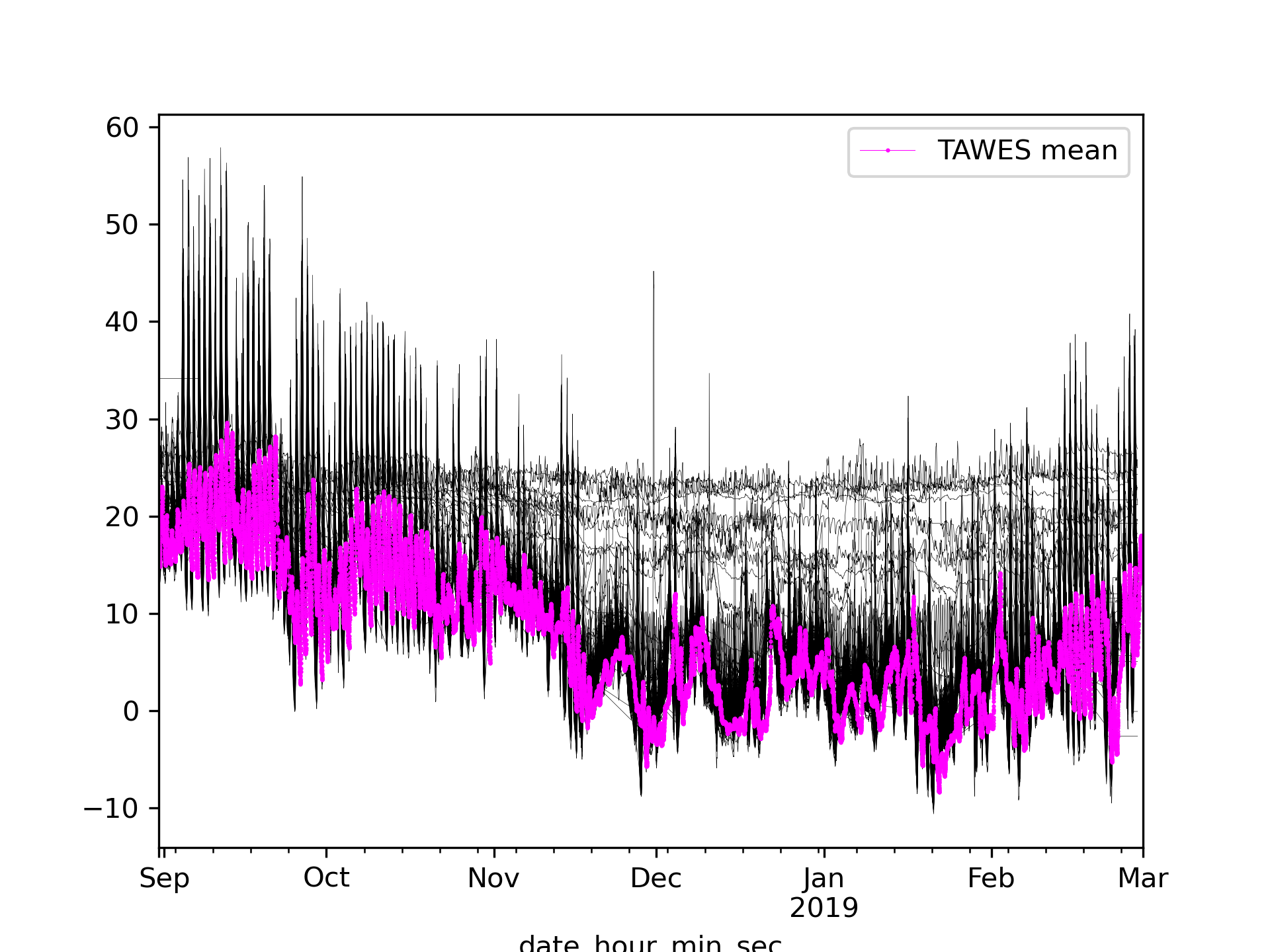}
\end{subfigure}
  \caption{Left: location of TAWES (blue triangle) and NetAtmo sites (pink dots) used in this study. Right: temperature measurements at NetAtmo sites (thin, black lines) and  average temperature  at the eight TAWES sites (magenta, linespoints).}
\end{figure}\label{fig:sites_measurements}

\vspace{-0.06cm}

\subsection{Outlier Detection Results}
The first step after the clustering has been completed is to identify the outliers and remove them from the result. As explained in Section \ref{alg:description}, outlier points have a very high likelihood of ending up alone in a cluster. Conversely, we can identify outliers by looking for clusters that contain very few data points. The maximum number of data points in such an "outlier cluster" can be defined by the user and changed as required. In the strictest case, only data points that are completely alone in their cluster would be labelled as outliers.

A good first step towards determining such a "cutoff value" is to look at the relative change of the number of data points in each cluster. If the data set contains clearly identifiable outliers, then in all likelihood there is going to be a considerable step between the sizes of the outlier clusters and the inlier clusters. The location of this step would give a logical point to place the cutoff value.\footnote{If no clearly visible step in cluster size between outliers and inliers can be found, it might be useful to vary the threshold value used for clustering and re-evaluate the results. As described in more detail in Section \ref{alg:description}, both an extremely big and an extremely small threshold can be a reason why outliers do not show up clearly in the result, even though there might be a clear distinction in the data set.}

\begin{figure}
\begin{subfigure}[t]{.24\textwidth}
  \centering
  \includegraphics[width=0.99\linewidth]{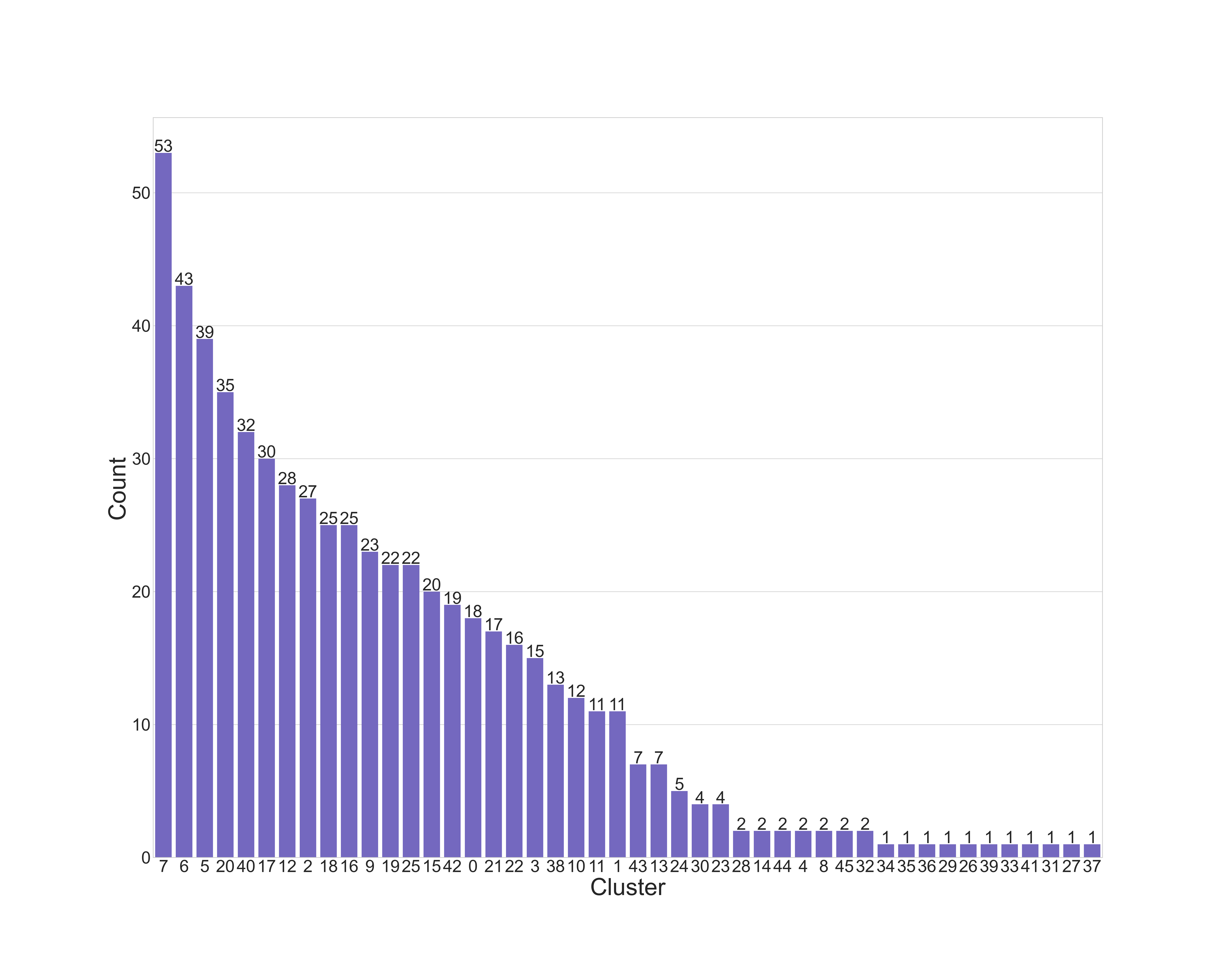}
\end{subfigure}
\begin{subfigure}[t]{.24\textwidth}
  \centering
     \includegraphics[width=0.99\linewidth]{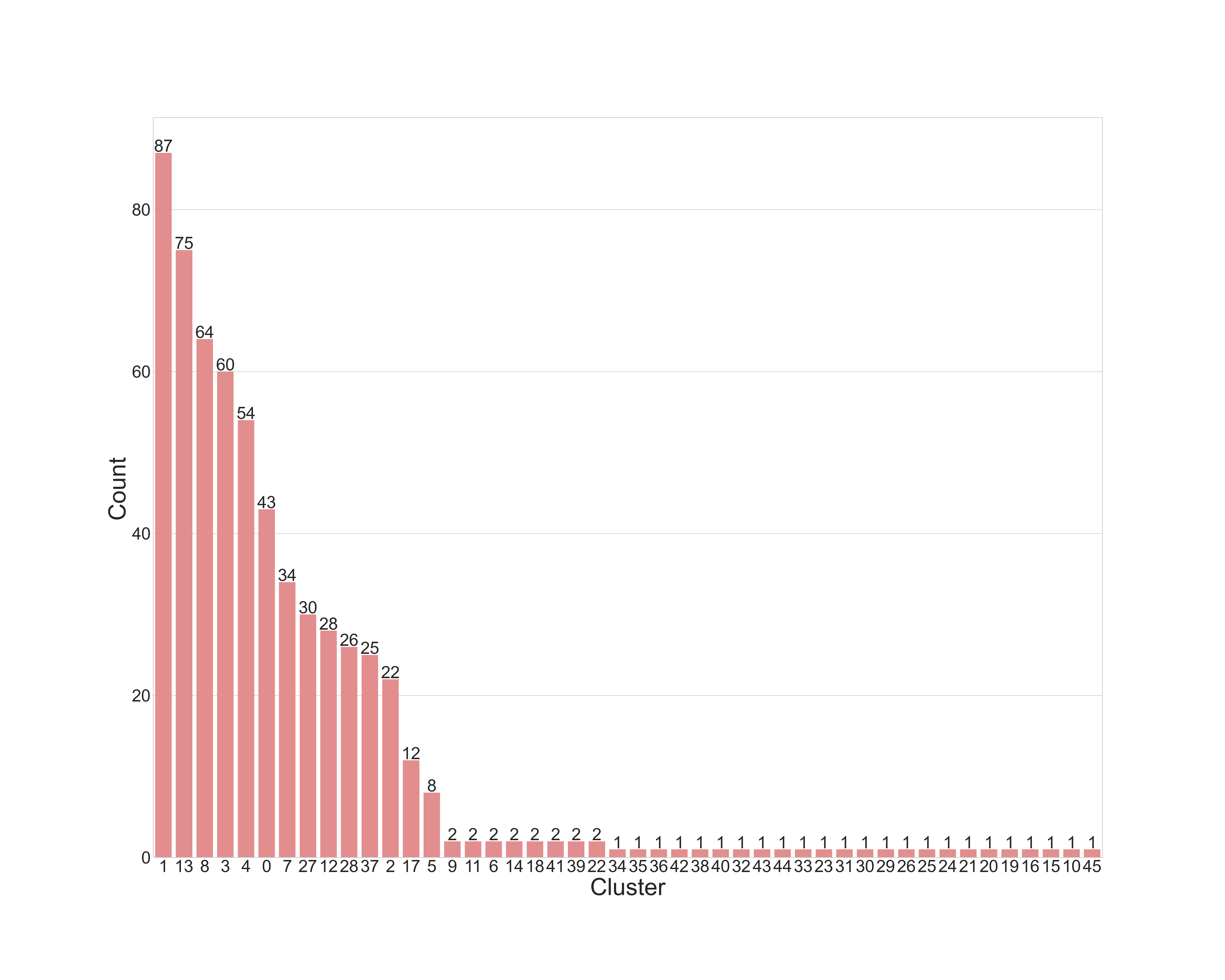}
\end{subfigure}
  \caption{Number of data points in each cluster for  BIRCH  (left)  and   AWT algorithm (right).}
\end{figure}\label{fig:cluster_sizes_barplot}

Figure 7 
shows the number of data points in each cluster of  both the BIRCH (Fig. 7 
left) and the AWT (Fig. 7 
right) clustering / outlier detection result, ordered by cluster size. The expected large step  in AWT can be found between the clusters with the IDs 5 and 9. At this point, the cluster size  in AWT decreases by a factor of 4 (from eight points in cluster 5 to two points in cluster 9), while at every other step the cluster size decreases by a factor of 2 or less. Therefore, this step is the logical choice for the distinction between inliers and outliers.  For the BIRCH algorithm this steep decrease in cluster sizes is not found as the number of member gradually decreases. Here, the delineation was made starting with cluster sizes of two or less. The results, based on the number of members within the clusters, shows that the AWT algorithm has a higher number of outlier clusters than BIRCH. Furthermore, AWT shows steep declines in cluster members, thus grouping more members into fewer clusters in contrast to BIRCH. 

This decision leads to 14 clusters being classified as inlier clusters, and 32 clusters being identified as outlier clusters  for the AWT algorthm and 28 inlier and 18 outlier cluster for the BIRCH algorithm. In terms of data points, 40 of the 608 points in the data set were classified as outliers  in AWT and 25 in BIRCH.
The choice of cutoff value is further confirmed by visual inspection of the result. Fig. 1  and  Fig. 2 in  the supplementary material under  \url{https://t1p.de/9qh18}
show the mean of the time series in each cluster, divided into inlier and outlier clusters for both algorithms BIRCH (top plots) and AWT (bottom plots). The inlier time series in Fig. 1 have a predictable form. Their overall trend goes downward for most of the time before it starts increasing again slightly, coinciding with the general temperature developments in the climatology for the period between September and February. In addition, they also exhibit short-term fluctuations that are due to the temperature differences between night and day. The outlier data in Fig. 2, on the other hand, deviates considerably from this expectation.  For some clusters/sites one could conclude that their location has been shifted during the period considered, whereas others remain indoors/at locations with very little temperature fluctuation
which do not even reproduce the climatological cycle of Vienna.

Inspecting the results also shows that for this data set some parameter tuning in both AWT and BIRCH might be
needed. Cluster 41 in AWT, as an example, contains a station which here is clearly marked as outlier, whereas the
BIRCH algorithm marks it as inlier in cluster 0. Or one of the sites in the AWT outlier cluster 33 is an inlier
in BIRCH in cluster 23. On the other hand some stations are marked as outliers in BIRCH which can be considered
as outliers in terms of temperature values but are seen as inliers in AWT. We can also observe that some sites might
have minor changes which mark them as outlier, although for the remaining period the data would fit into one of the
inlier clusters. This could be avoided by using different time slices in both algorithms or, additionally, a break point
detection. Although both algorithms are able to detect a large number of outliers with AWT detecting more correct
outliers for the time span considered in this study, both a thorough re-evaluation of the clustering parameters as well
as a break point detection for those sites which partially are in accordance with the observations are required steps
for future offline and online applications.

\subsection{Clustering Results}\label{ss-clusteringresults}

Once the outliers have been removed, it is possible to analyse and interpret the remaining data clusters.

\begin{figure*}
\begin{center}
\begin{subfigure}[t]{.38\textwidth}
  \centering
  \includegraphics[width=\linewidth]{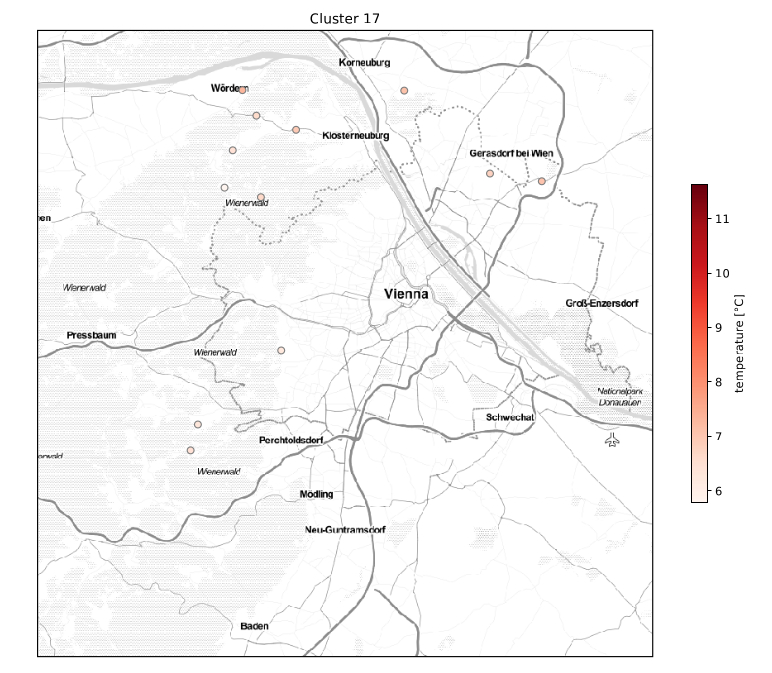}
    \caption{Cluster 17:   stations located in less urbanized \\areas near or  outside the  Vienna city borders.}
    \label{fig:cluster:outside-vienna}
\end{subfigure}%
\hspace{0.5cm}
\begin{subfigure}[t]{.38\textwidth}
  \centering
   \includegraphics[width=\linewidth]{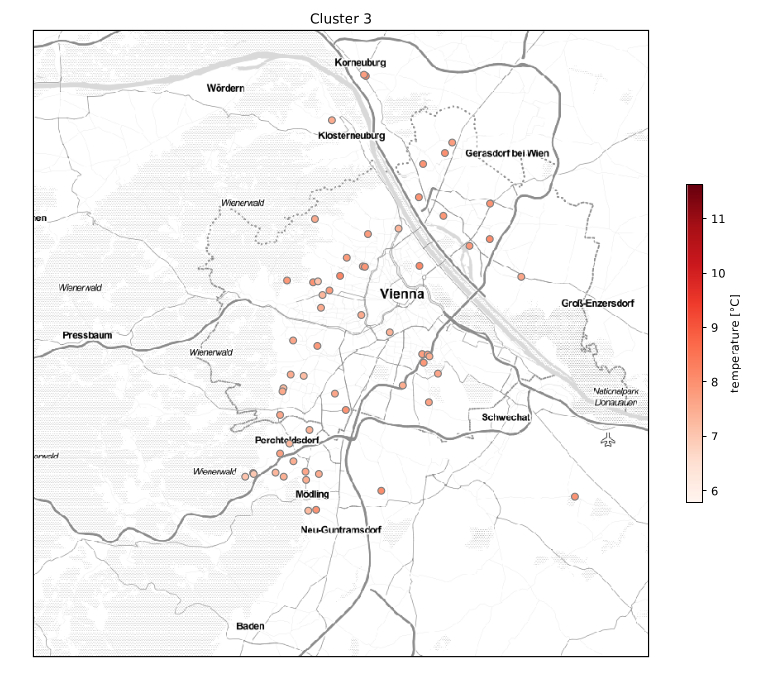}
  \caption{ Cluster 3: mostly stations located in the outer Vienna districts, specifically in the West.}
  \label{fig:cluster:outer-districts}
\end{subfigure}
\\
\hspace{0.25cm}
\begin{subfigure}[t]{.38\textwidth}
  \centering
  \includegraphics[width=\linewidth]{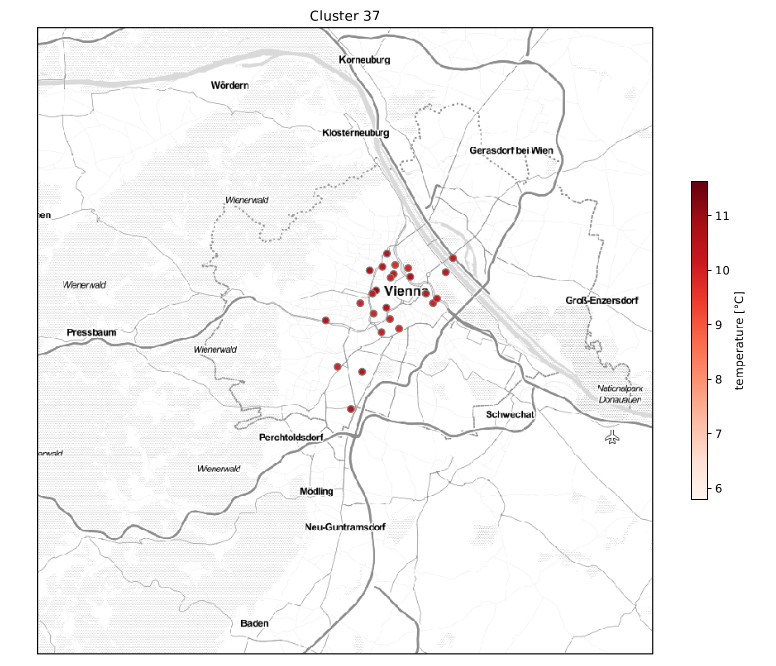}
    \caption{Cluster 37:  almost exclusively  stations  \\inside the G\"urtel, one  main traffic road \\in Vienna.}
    \label{fig:cluster:inner-districts}
\end{subfigure}
\hspace{-0.5cm}
\begin{subfigure}[t]{.45\textwidth}
  \centering
  \includegraphics[width=\linewidth]{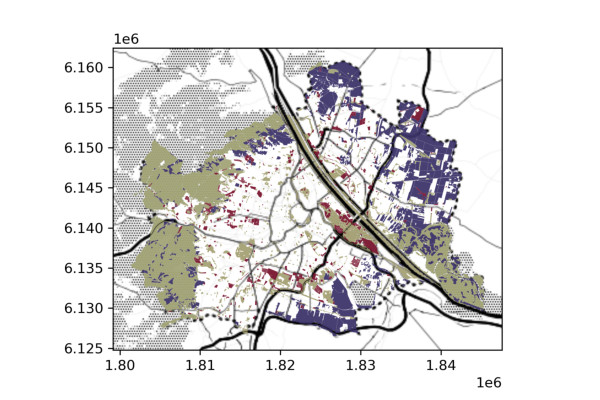}
    \caption{Vienna: green urban areas = red and green, agricult. sites including vineyards = blue for agricult. sites  incl.  forests in and around Vienna (gray).}
    \label{fig:cluster:greenvienna}
\end{subfigure}

\caption{Example clusters in the map of Vienna. The colour of a station indicates its average temperature in the observed time period. In  lower right the different green urban areas (parks,  recreation areas, vineyards, etc.) are from  Viennese  land-use atlas.}
\label{fig:clusters}
\end{center}
\end{figure*}

While Vienna is not an extremely large city, it is big enough that geographical concentration of clusters in different parts of the city can be expected. Furthermore, due to the topography with the outskirts of the Alps in the western and southern part of Vienna, differences between the cluster temperatures are expected.
When plotting the resulting clusters on a map of Vienna, we find that these effects are indeed present in our data. To illustrate this, Figure \ref{fig:clusters} shows the stations from three example clusters. In these clusters, the effects described above are particularly pronounced: The geographical coherence of stations inside the same cluster can be clearly seen, as well as a general increase in temperature towards the city center.

A station's temperature is also likely to be influenced by the land usage in the area where it is positioned (for example, the density of constructions or the amount of greenery that can be found in its surroundings). Figure \ref{fig:cluster:greenvienna} shows land usage types in different areas of Vienna.
Given the examples in Figure \ref{fig:clusters} we can see that sites located in similar environments are grouped into one cluster.


\subsection{Reducing the Wavelet Resolution}

While it was possible to cluster this data set at the full available wavelet resolution, it is still desirable to examine the effect that reducing the wavelet resolution has on the quality of the result. An easy metric to estimate this is the \emph{Normalized Mutual Information} (NMI), which gives an indication of the similarity of two clustering assignments
\cite{NMI}. The NMI returns a value between 0 and 1, where 1 indicates that the cluster assignments match exactly, and 0 means that there is no correlation between the assignments.
We used the NMI to determine the similarity between the result at full wavelet resolution and results achieved with reduced wavelet resolution. For these experiments we dropped up to three wavelet levels. Since dropping one wavelet level cuts the amount of data approximately in half, this is equivalent to reducing the required memory to up to $1/8$ of the size of the original data set.
In addition to examining the effect on our standard input settings, we also conducted experiments where the clustering feature tree is created using five instead of three wavelet levels. The threshold value was adjusted to 1.51 to create the same number of clusters as in the original experiments, allowing for better comparability. \footnote{The number of created clusters for any given settings is the same, regardless of  whether wavelet levels have been dropped. The number is determined while  constructing  the clustering feature tree, which only uses a fixed subset of the available levels and  not  the total number of available levels.}
These results are summarized in Table \ref{tab:resolution}.

As expected, the result of the experiments with reduced wavelet resolution tends to deviate stronger from the full-resolution result as more wavelet levels are dropped.
With the first set of experiments, the NMI is slightly below 0.9 even if only one wavelet level is dropped. This indicates a deviation from the original result that is not dramatic, but still larger than one could wish for.
The second set of experiments, however, exhibits much more stable behaviour: Dropping one level leads to an NMI of above $0.98$, a deviation that would hardly be noticeable in the final result. The only experiment in this set where the NMI falls below $0.9$ is when three levels are dropped, which is quite acceptable w.r.t the massive reduction of the data size in this case.
Evidently, the negative effects of reduced wavelet resolution can be counterbalanced at least to a certain extent by increasing the number of wavelet levels used for the creation of the clustering feature tree.

\begin{table}[ht]
    \centering
    \begin{tabular}{r|r|r|r}
        CF Tree Levels & \multicolumn{3}{c}{max resolution}  \\
        & 12 & 11 & 10 \\
        \hline
        3 & 0.890 & 0.898 & 0.846 \\
        5 & 0.984 & 0.946 & 0.890 \\
    \end{tabular}
    \caption{NMI between cluster assignments with full wavelet resolution (WR) and cluster assignments with reduced WR. \emph{CF Tree Levels} = the number of wavelet levels used to build the clustering feature; \emph{max resolution} = the maximum number of wavelet levels available for I-Kmeans.}
    \label{tab:resolution}
\end{table}

\section{Conclusion} \label{sec:conclusion}

We presented the \algname algorithm, a clustering algorithm for meteorological time series data. It is a variation of the BIRCH and I-Kmeans algorithms that reinterprets the threshold parameter as a measure of the desired clustering granularity. While it has already been shown that a combination of I-Kmeans and BIRCH algorithms can be used to obtain good initial guesses of the cluster centers, our approach also resolves the issue of determining the $k$ parameter: The number of clusters in the output is automatically determined based on the threshold value. Furthermore, our algorithm tolerates and detects the presence of outliers in the input data. For experimental evaluation, we used our algorithm to cluster crowd-sourced meteorological data from Vienna. Our experiments show that the results achieved by this algorithm are very plausible in terms of outlier detection and clustering and indicate a high clustering quality. Furthermore, one can detect different urban land-use classes in the representation of the sites in the clusters. This enables future application of the clustering algorithm for tasks such as surface temperature interpolation of privately owned non-standard observation sites, data assimilation in NWP models or clustering of sites to enable statistical and machine learning models to train on larger data sets.\\
While we presented an experimental evaluation of \algname and a comparison with the BIRCH algorithm, a more detailed analysis of \algname is yet required to fully understand all of its capabilities and limitations in different settings. Avenues for future research include a thorough investigation of the effects of the threshold parameter, as well as a comparison of \algname with additional existing clustering algorithms.\\
Our code in  C++  with Python bindings is publicly available under
 \url{https://t1p.de/9qh18}.

\section{Supplementary Material}
Here we present additional figures to Section~III.B.
Each figure shows all the data from one cluster in black, with the mean of the TAWES observation data in magenta. The data in the outlier clusters differs much more strongly from the TAWES mean than the data in the inlier clusters.

\begin{figure}
  \centering
  \hspace{0.3cm} \includegraphics[width=0.85\linewidth]{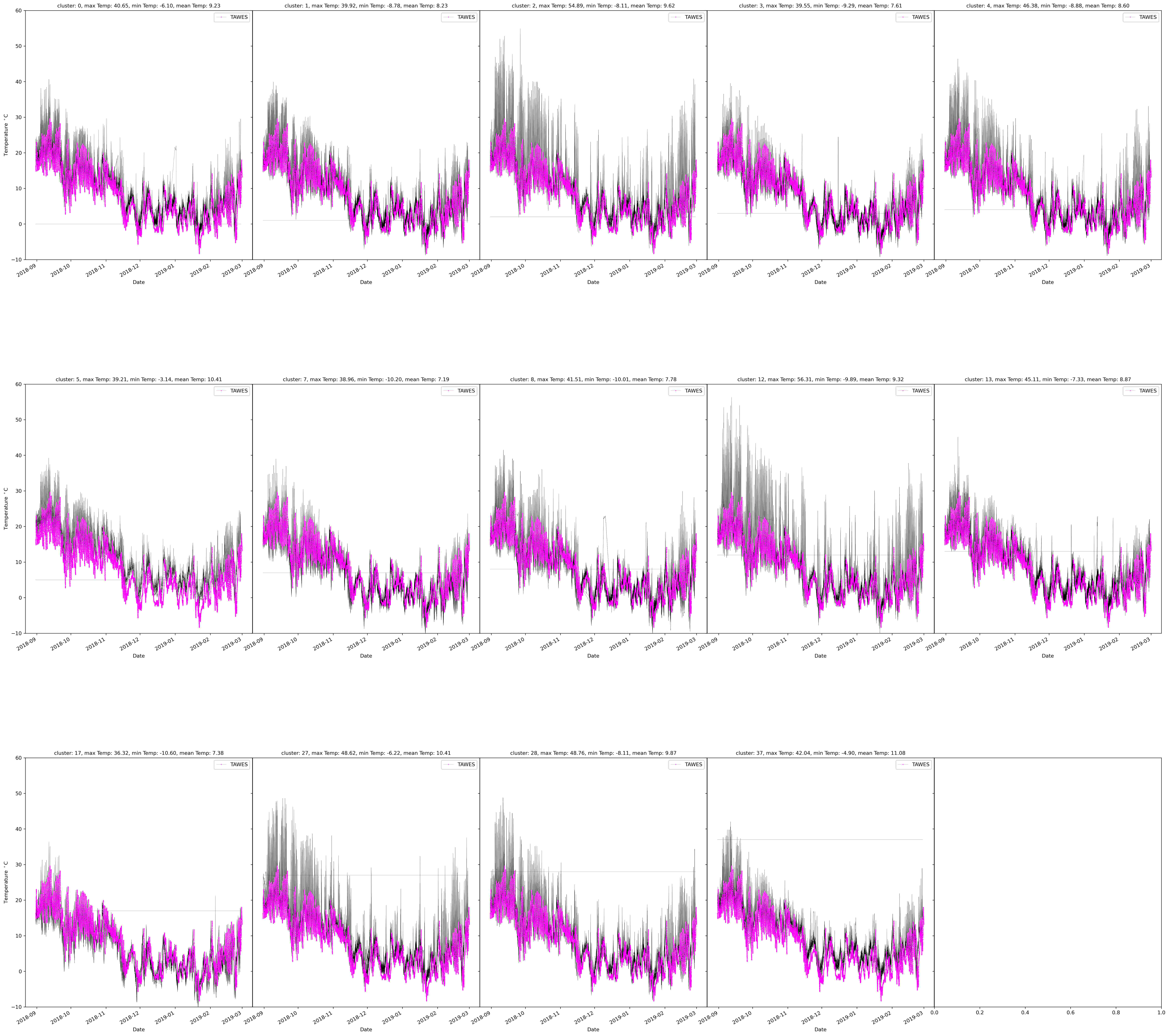}\\
  \vspace{0.4cm}
    \caption{Time series of the stations contained in the inlier clusters for AWT.}
    \label{fig:11}
    \end{figure}
\begin{figure}
  \centering
\hspace{-0.3cm}
   \includegraphics[width=1.1\linewidth]{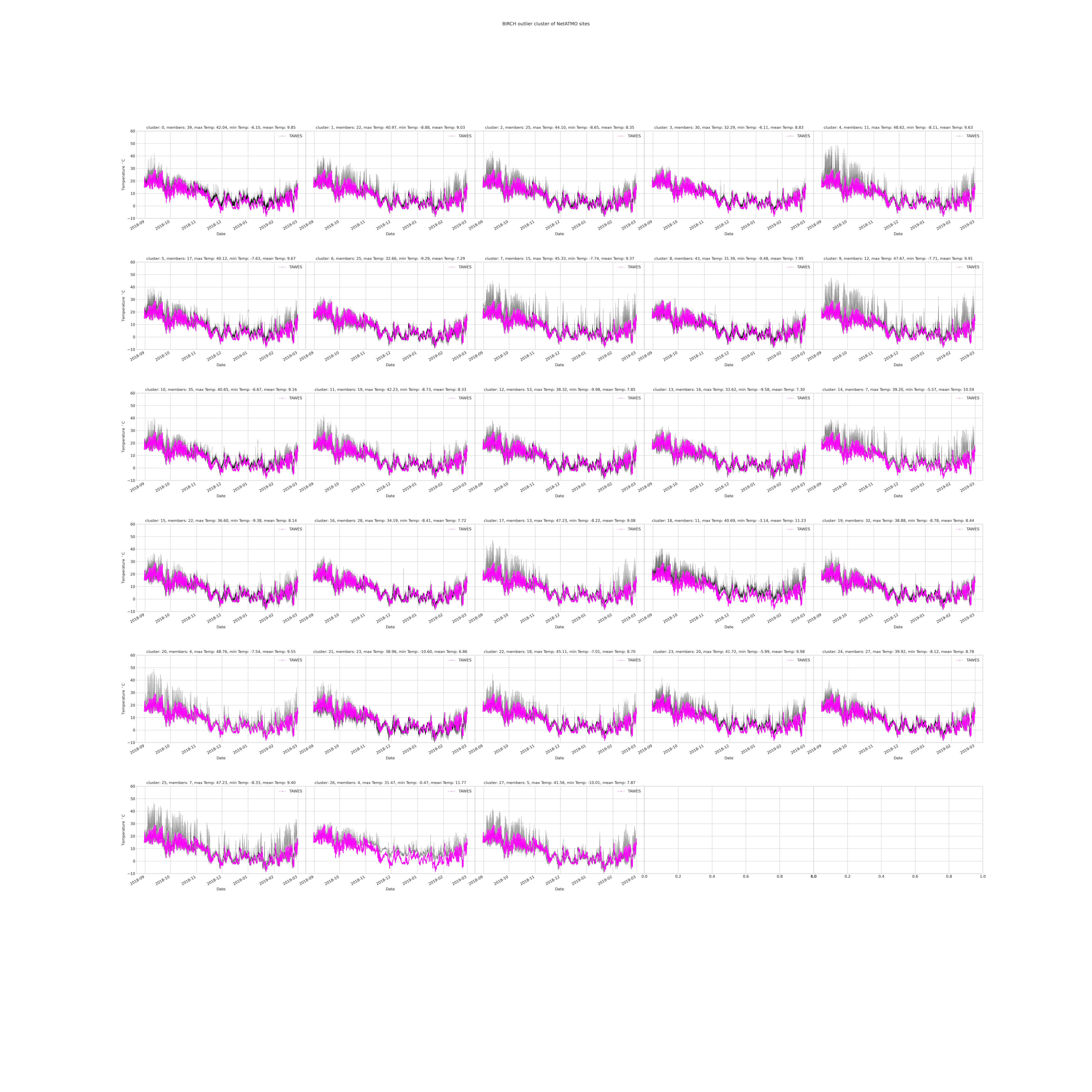}\\
  \vspace{-0.2cm}
  \caption{Time series of the stations contained in the inlier clusters for BIRCH.}
  \label{fig:12}
\end{figure}

\begin{figure}
  \centering
  \includegraphics[width=0.85\linewidth]{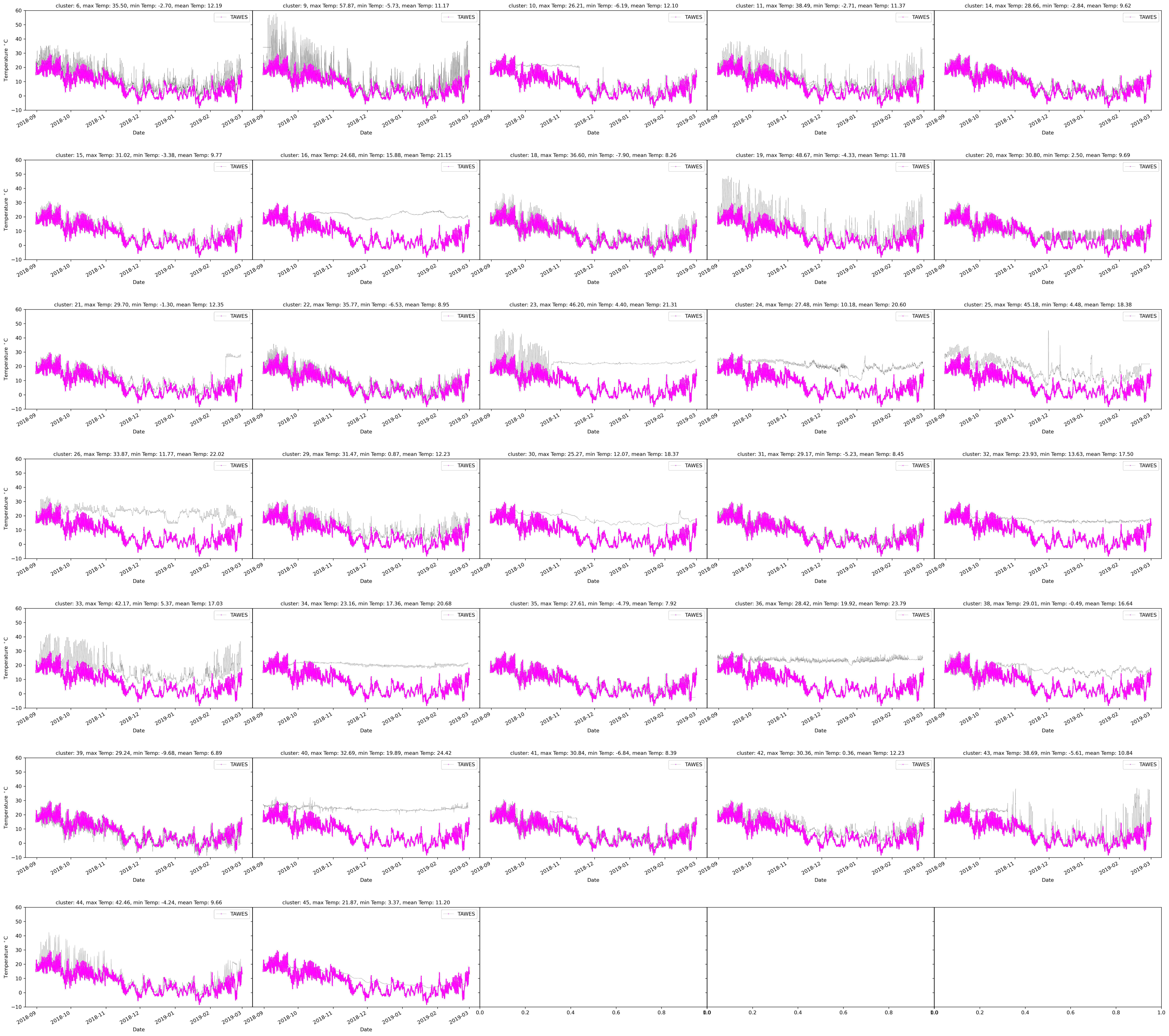}\\
  \vspace{0.2cm}
    \caption{Time series of the stations contained in the outlier clusters for AWT.}
    \label{fig:13}
\end{figure}

\begin{figure}
  \centering
\hspace{-0.6cm}  \includegraphics[width=0.95\linewidth]{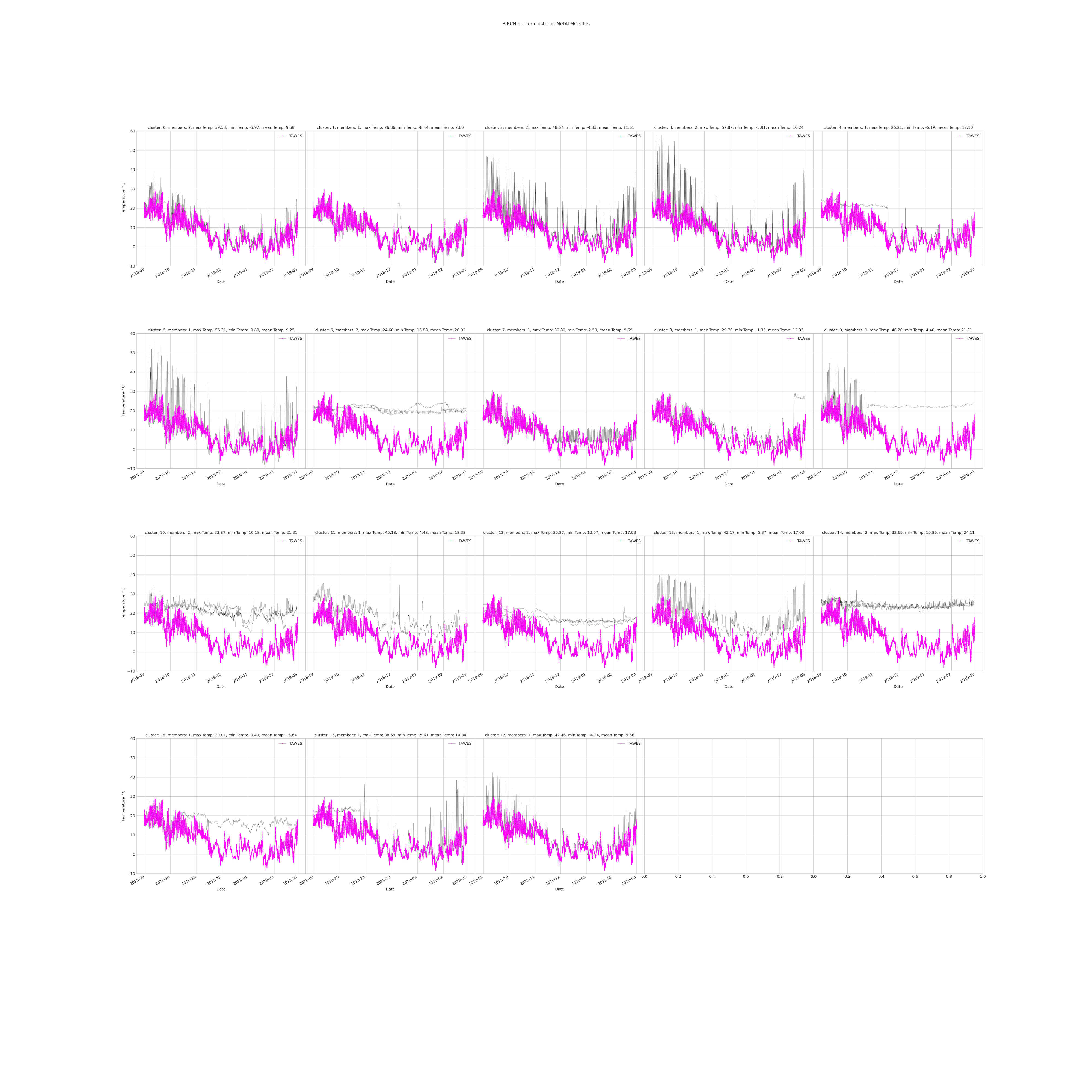}\\
\vspace{-0.3cm}
    \caption{Time series of the stations contained in the outlier clusters for BIRCH.}
    \label{fig:14}
\end{figure}




\end{document}